# A Machine Learning Approach for Early Detection of Fish Diseases by Analyzing Water Quality

Al-Akhir Nayan[1,*], Joyeta Saha[1], Ahamad Nokib Mozumder[1], Khan Raqib Mahmud[2], Abul Kalam Al Azad[2] and Muhammad Golam Kibria[2]

[1]*Department of Computer Science & Engineering, European University of Bangladesh, Dhaka, Bangladesh*
[2]*Department of Computer Science & Engineering, University of Liberal Arts Bangladesh, Dhaka, Bangladesh*

([*]Corresponding author's e-mail: asquiren@gmail.com)

**Abstract**

Early detection of fish diseases and identifying the underlying causes are crucial for farmers to take necessary steps to mitigate the potential outbreak and thus to avert financial losses with apparent negative implications to the national economy. Typically, fish diseases are caused by viruses and bacteria; according to biochemical studies, the presence of certain bacteria and viruses may affect the level of pH, DO, BOD, COD, TSS, TDS, EC, $PO_4^{3-}$, $NO_3$-N, and $NH_3$-N in water, resulting in the death of fishes. Besides, natural processes, e.g., photosynthesis, respiration, and decomposition, also contribute to the alteration of water quality that adversely affects fish health. Being motivated by the recent successes of machine learning techniques, a state-of-art machine learning algorithm has been adopted in this paper to detect and predict the degradation of water quality timely and accurately. Thus, it helps to take preemptive steps against potential fish diseases. The experimental results show high accuracy in detecting fish diseases specific to water quality based on the algorithm with real datasets.

**Introduction**

Fishes account for approximately 15 % of the animal protein intake of the human population globally [1]. In countries like Bangladesh, fishes provide as high as 60 % of animal protein to the populace. Economic valuation fishes contribute approximately 3.6 % to the national GDP, weighing nearly 25 % of the entire agricultural GDP [2]. Furthermore, this section employs about 11 % of the total population in Bangladesh on a full-time and part-time basis [2]. Despite being the very vibrant economic sector for the country, one major threat to the fish farmers is fish diseases which eventually put a significant constraint on the economic progress and severely strain the expansion of aquaculture and fish farming [3]. Fish culture faces severe threats from waterborne pathogens, such as bacteria and viruses, responsible mainly for mass mortality and poor health. It is, therefore, imperative to monitor the purity of the water habitat to detect fish diseases timely and accurately.

Fish perform all their physical activities underwater; fish dependents on the water for breathing, feeding, reproducing, and growth. When the water quality of the habitat deteriorates, it becomes unfavorable for fish to live. Water quality depends on specific parameters, and when the parameters change, the quality deteriorates. As a result, the health of fishes is threatened by the compromise of their immune system, critically leading them to be vulnerable to harmful pathogens [4]. Oxygen plays a pivotal role in maintaining life underwater. When oxygen level goes below the preferable range, fish species' physiological and physical growth are hampered. Decreasing oxygen levels underwater causes acidosis and nephron calcinosis. It obstructs the development of granulomas in many internal organs of fish [5,6].

Moreover, recirculated water contains high pH, which turns up the ammonia level in the water. A high level of ammonia causes harm to the gills and liver of fishes. Depending on the level of saturation and the time of exposure, Gas supersaturation of the water can result in the gas bubble disease. The leading cause of the illness is bubbles in the eyes, skin, and gills [7]. Degraded water quality causes pollution that creates serious problems to fish; necrotic alteration, papilloma, degenerative, and fin erosion result from water pollution. As a result, the fish body gets abnormal growth, and farmers do not get optimum production. If the farmers could identify them early, they could be solved quickly. Machine learning techniques and artificial intelligence algorithms have been used extensively and successfully in classification and decision-making problems [8,9]. Intelligent algorithms can learn from the system's parameter space the correct desired classification based on a real dataset and infer very accurate deviation from the selected sets of configuration, which may be exploited for decision-making. A hierarchical architecture to the algorithm ensures higher performance from learning. Machine learning technique, such as Gradient boosting, exploits decision-tree based in-built hierarchical structure based on regression algorithm to classify complex problems and help decision-making automation [10,11]. Inspired by the successes of such a technique in various challenging situations, we have employed the technique in this study to predict fish diseases by solving classification problems from real datasets composed of desired parameters of water purity. Our approach is to the development of an automated fish disease detection algorithm, and employ it in the decision-making process may be summarized in the following sequential steps:

Step 1: Take a sample of water to identify the water quality.
Step 2: Predicting the water quality using a machine learning algorithm. We have already collected and prepared a dataset and trained our algorithm to predict probable fish diseases based on the water quality parameters.
Step 3: Analyzing the disease and identifying.
Step 4: Making wise decisions to minimize harm to the fish farm and ensure healthy habitat.

The paper is arranged as follows: the literature review is conducted in the second section, followed by an in-depth look into the model's proposed machine learning algorithm and parameters in section 3. The dataset is discussed in section 4 and the preparation, processing, and implementation in the model. Finally, the experimental results are discussed in section 5, followed by concluding remarks.

**Related works**

In 2012, Chemical Oxygen Demand (COD), Total Suspended Solids (TSS), and Oil & Grease (O&G) were monitored using a multi-sensor water quality monitoring system. An improved boosting method was used for suppressing quality-relevant variables by applying smaller weights and constructing models based on weighted variables for wastewater quality predictions. The observing framework was tried in the field with good outcomes, hiding this method's capability to check water quality online [12]. Another study took place in 2014 to measure the pollution of the South Korean Ocean using machine learning techniques. They collected 63 samples from 2011 to 2012. The study successfully discussed the distribution of water quality parameters analyzing the Geostationary Ocean Color Imager (GOCI) images [13]. In 2012, a study by S. Shah was carried out on twenty different locations of Kerala, India. The study showed that most water samples in the region were suitable for agriculture purposes. A very easy pretreatment was also necessary to make the water ideal for underwater lives [14]. N. Karlar collected water samples from 10 villages and analyzed them. He analyzed Physico-chemical variables like TH, Temp, F—, PH, Cl—,TDS, Ca2+, Mg2 +, alkalinity, and SO4 2-. The WQI of these specimens ranged from 40.67 to 69.59, showed that block surface water was needed to be processed before use [15].

In 2013, Usha researched the water quality rating and health of urban water bodies in Bilari town. Water samples were obtained from ten separate locations in the three months of January, June, and September 2011. Examination of surface water was done for various physico-synthetic parameters. WQI indicated that water contamination was rising day by day [16]. A. B. Frontier worked on his research. To analyze the Physico-chemical properties of water, samples were obtained from selected villages in Nasik district of Kalwan Tahsil. Water tests were gathered haphazardly from 5 distinct locations. The result shows that physico-compound parameters of gathered water tests are inside good cutoff in the study zone. So it could be said that water condition was favorable for the lives underwater [17]. N. Bagde researched the assessment of water quality and its effects with specific reference to Chhindwara district. Water samples from 5 blocks of the infected area were obtained and examined for physico-chemical variables such as alkalinity, pH, total hardness, electrical conductivity, and fluoride ion. Many water samples showed a greater concentration of fluoride ions and greater turbidity. Most of the samples needed chemical treatment to make them suitable for the living body. Adeyemi et al. researched the chemical impact of leachate on the consistency of Nigeria's surrounding water. The analysis showed that the BOD and COD of the leachate-contaminated water samples were more excellent in the dry season than in the rainy season. Bacteria were more in the rainy season than the dry season. Consequently, the data of this study showed that the use of leachate-contaminated water was harmful and should be prevented [18].

To evaluate the water quality index, physicao-chemical experiments like pH, dissolved oxygen chloride, temperature, total alkalinity, total dissolved solids, calcium & sulfate, magnesium hardness, phosphate, and bore well water nitrate were conducted in 20 villages Gondpipri and some of its interior Adivasi areas during 2010. The findings show that pH & TDS is within the same limits that WHO & ISI provide. Likewise, calcium, Mg, Sulphate are within permissible limits. But the phosphate levels, nitrates are higher values than prescribed values are harmful to fish farming [19]. In another study, Physico-chemical parameters such as temperature, pH, sulfate, electrical conductivity, bicarbonate, sodium, calcium, potassium, magnesium, iron, silica, chloride, nitrate, dissolved oxygen, phosphate, physical demand for oxygen & chemical demand for oxygen were monitored. From 2017 February to 2018 March, samples of those parameters were collected and analyzed from 8 separate reservoir locations. This analysis shows a frequent variety of convergence of Physico-chemical parameters, and some of the parameters are past passable breakpoints, showing water quality degradation due to contamination [20].

Most of the studies indicate the change in water quality due to the transformation of the chemical components of water. Inspiration drawn from these studies has led to the current investigation on water quality governed by the parameters based on the chemical composition of water. And the study further links this degradation of water to the emergence of various fish diseases.

**Research method**

    **Gradient boosting technique**

Gradient boost is a sophisticated machine learning tool used for solving regression and classification problems in complex decision-making tasks. The model is comprised of computational stages made up of decision trees. The model executes a given classification task based on minimizing a differentiable loss function in every computational step of the architecture. The stages are coupled within the learning rule with a weak learning rate in an additive formalism. The underlying feature of this functional architecture is that as the model learns from every successive computation, the performance gets better gradually as it learns to minimize its mistake by gauging the loss function. Thus, the model learns the complex inter-relationship among the parameters and variables in small steps, as if the algorithm matures gradually with every model generation. The working procedure of the model is shown in Figure **1**.

Algorithm 1: Gradient boost algorithm

**Inputs:**

- Data input $(x, y)_{i=1}^{N}$
- Iterations number $M$
- The loss-function $\Psi(y, f)$
- The base-learner model $h(x, \theta)$

**Algorithm:**

1. Initialize $\widehat{f_0}$ with constant.
2. For $t = 1$ to $M$ **do**
3.    Negative gradient computation $g_t(x)$
4.    Fit a new base learner function $h(x, \theta_t)$
5.    Find the best gradient decent step-size $\rho_t$:

$$\rho_t = \arg\min_\rho \sum_{i=1}^{N} \Psi[y_i, \widehat{f_{t-1}}(x_i) + \rho h(x_i, \theta_t)]$$

6.    Update the function estimate:

$$\widehat{f_t} \leftarrow \widehat{f_{t-1}} + \rho_t h(x_i, \theta_t)$$

**END**

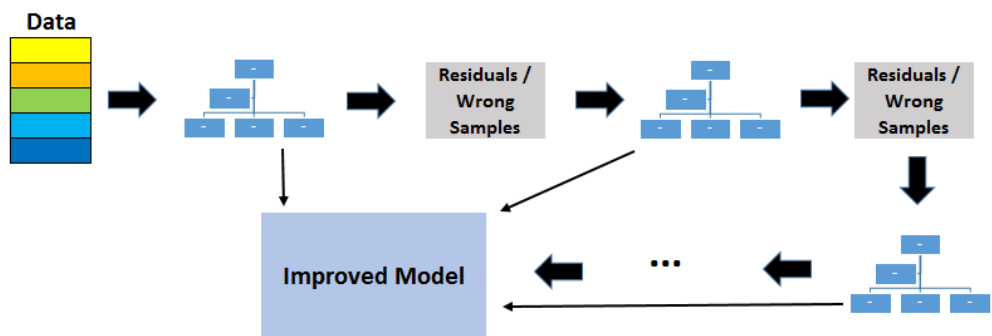

**Figure 1** Working procedure of gradient boosting.

The numeric values of the model parameters used in the simulation are shown in **Table 1**.

**Table 1** Numeric value of model parameters.

| Name of parameter | Value |
|---|---|
| Minimum sample split | 200 |
| Learning rate | 0.1 |
| Max depth | 8 |
| Minimum sample leaf | 30 |

**Model parameters and description**

**Table 2** below shows the computational types of the parameters used in the model to calculate and predict water quality.

**Table 2** Model parameters.

| Parameter name | Type | Unit |
|---|---|---|
| STATION CODE | object | |
| LOCATIONS | object | |
| STATE | object | |
| Temp | float64 | |
| D. O. | float64 | mg/L |
| pH | float64 | |
| CONDUCTIVITY | object | mhos/cm |
| B.O.D. | float64 | mg/L |
| NITRATENAN N+ NITRITENANN | float64 | mg/L |
| FECAL COLIFORM | object | MPN/100 mL |
| TOTAL COLIFORM | float64 | (MPN/100mL) Mean |
| Year | Int64 | |

Float64 has been taken because of storing fraction values for the parameters. In the table, the parameter Temp indicates the temperature of the water. The station code, location, and state provide information about the site where the sample data has been collected. D.O, which stands for Dissolved Oxygen, indicates the free and non-compound oxygen level in the water. It is an essential parameter in assessing water quality due to its effect on the species within a water habitat. A too high or too low level of D.O. can damage aquatic life and affect water quality [21].

The pH indicates the amount of acid/basic compound in the water. Water pH regulates the solubility and biological quality of contaminants such as nutrients and heavy metals. In the case of heavy metals, toxicity is determined by the degree to which they are soluble. Conductivity measures the capability of passing an electric current through water [22,23]

Biochemical Oxygen Demand (BOD) measures the amount of dissolved oxygen used by aerobic microorganisms when the organic matter is decomposed in water. It offers an index for evaluating the impact of wastewater discharged on the recipient area. A high BOD value indicates many available organic compounds for bacteria that consume oxygen [24].

Nitrate is formed by combining oxygen or ozone with nitrogen. Nitrogen is helpful for all living bodies. But a higher level of nitrate is harmful to all the live bodies underwater. The term Coliform indicates bacteria that are present in animals, including humans. Coliform does not cause diseases, but some coliform like E. coli can cause serious harm to the living body [25,26].

**Calculations**

Water quality is determined by a variable WQI constructed based on npH, NBDO, NEC, NNA, WPH, WDO, WBDO, WEC, WNA and WCO. The variable is used to perform mathematical calculations, as shown in **Table 3**.

**Table 3** Calculation of chemical components in water.

| Calculation name | Component range | Calculated value |
|---|---|---|
| **Calculation of npH** | **PH range** | **npH value** |
| | 7 to 8.5 | 100 |
| | 8.5 to 8.6 or 6.8 to 6.9 | 80 |
| | 8.6 to 8.8 or 6.7 to 6.8 | 60 |
| | 8.8 to 9 or 6.5 to 6.7 | 40 |
| | Else | 0 |
| **Calculation of nco** | **TC range** | **Nco value** |
| | 0 to 5 | 100 |
| | 5 to 50 | 80 |
| | 50 to 500 | 60 |
| | 500 to 1000 | 40 |
| | Else | 0 |
| **Calculation of nec** | **CO range** | **Nec value** |
| | 0 to 75 | 100 |
| | 75 to 150 | 80 |
| | 150 to 225 | 60 |
| | 225 to 300 | 40 |
| | Else | 0 |
| **Calculation of ndo** | **DO range** | **ndo value** |
| | 6+ | 100 |
| | 5.1 to 6 | 80 |
| | 4.1 to 5 | 60 |
| | 3 to 4 | 40 |
| | Else | 0 |
| **Calculation of nbdo** | **BOD range** | **Nbdo value** |
| | 0 to 3 | 100 |
| | 3 to 6 | 80 |
| | 6 to 80 | 60 |
| | 80 to 125 | 40 |
| | Else | 0 |
| **Calculation of nna** | **NA range** | **Nna value** |
| | 0 to 20 | 100 |
| | 20 to 50 | 80 |
| | 50 to 100 | 60 |
| | 100 to 200 | 40 |
| | Else | 0 |

**Now,**

$wph = npH * 0.165$

$wdo = ndo * 0.281$

$wbdo = nbdo * 0.234$

$wec = nec * 0.009$

$wna = nna * 0.028$

$wco = nco * 0.281$

$wqi = wph + wdo + wbdo + wec + wna + wco$

The value of the variable WQI measured as given is used in the learning algorithm to predict the water quality based on the constituting parameters; a particular value of the variable relates to a specific fish disease [27,28].

**Dataset collection and configuration**

A dataset was prepared for training the model. Approximately 2000 samples were collected from different places in Bangladesh. Before training, the data was preprocessed. **Table 4** provides information about the collected dataset. At the time of taking the sample, we collected other necessary information for our calculation process. All the terms used in **Table 4** have already been described in the method. According to medical science, we have found the probable disease and suggestions depending on the water quality range. **Table 5** provides a short description of the disease and the suggestions. We have trained our model with 63 conditions, their symptoms, and recommendations. 3D visual representation of filtered data has been shown in **Figure 2**, while **Figure 3** depicts the 2D representation of **Figure 2**. The X-axis of **Figure 2** represents the time in the year, and the Y-axis represents the water quality. We note from **Figures 2** and **3** that the water quality was better in 2017 than in other years.

**Table 4** Dataset configuration.

| Serial No | Station code | Locations | State | Temp | D.O. (mg/L) | pH | conductivity | B.O.D. | NITRATENAN N+ NITRITENANN (mg/L) | FECAL COLIFORM (MPN/100 mL) | Total COLIFORM (MPN/100mL) Mean | Month and year |
|---|---|---|---|---|---|---|---|---|---|---|---|---|
| 0 | 1207 | Dhanmondi Lake Area, Dhaka | Dhaka | 30.6 | 6.7 | 7.5 | 203 | 1.3 | 0.1 | 11 | 27 | 8-2019 |
| 1 | 1207 | Dhanmondi 27 Area, Dhaka | Dhaka | 29.8 | 5.7 | 7.2 | 189 | 2 | 0.2 | 4953 | 8391 | 8-2019 |
| 2 | 1208 | Mirpur Area, Dhaka | Dhaka | 29.5 | 6.9 | 6.9 | 179 | 1.7 | 0.1 | 3243 | 5330 | 8-2019 |
| 3 | 9210 | Doulotpur Area, Khulna | Khulna | 29.7 | 5.8 | 6.9 | 64 | 3.8 | 0.5 | 5382 | 8443 | 9-2019 |
| 4 | 9450 | Shyamnagar Area, Satkhira | Satkhira | 28.2 | 5.6 | 7.3 | 83 | 1.9 | 0.4 | 3428 | 5500 | 9-2019 |

**Table 5** Diseases with relevant reason and suggestion.

| Serial | Disease | Reason | Suggestion |
|---|---|---|---|
| 1 | Acid death | Low pH level | Use chemical to increase basic compound |
| 2 | Fungus in mouth | Caused from bacterium | Use of Penicillin |
| 3 | Growth process slow | Lack of protein | Protein synthesis |
| 4 | Tuberculosis | Driven by the bacterium Mycobacterium piscium | Destroy infected fish |
| 5 | Tail rot & Fin rot | Caused by the bacteria Aeromonas | Use $CuSO_4$ |
| … | … | … | … |
| 58 | Alkaline death | Decrease of hydroxyl ion in water | Use chemical to increase acid compound in water |

| Serial | Disease | Reason | Suggestion |
|---|---|---|---|
| 59 | Ulcer | caused by bacteria, Haemophilus | $CuSO_4$ for 1 min for a period of 3 to 4 days |
| 60 | Ichthyosporidium | Caused by fungus | Add phenoxethol to food |
| 61 | Velvet or rust | Infection in gill and velvet by virus | Copper at 0.2 mg/L |
| 62 | Nematoda | Nematodes (threadworms) infect | soak the food in para-chloro-meta-xylenol |
| 63 | No production | Bacteria attack | Minimize acidity by using soda lime |

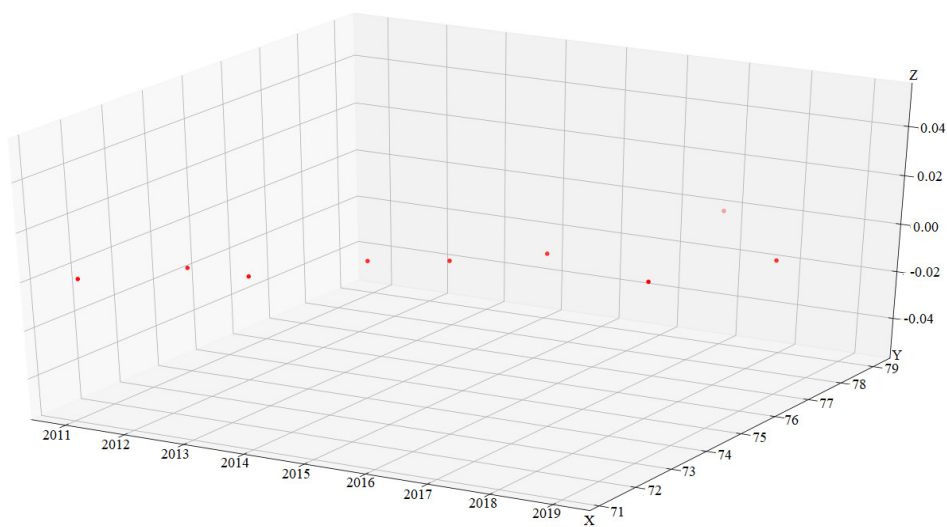

**Figure 2** 3D Visualization of filtered data.

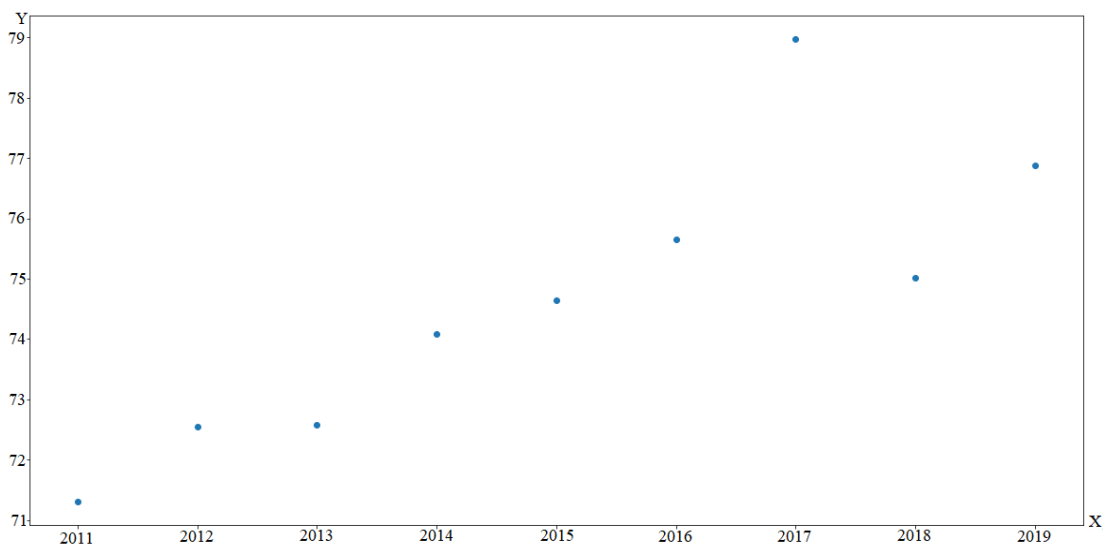

**Figure 3** Planar scatter plot of data points.

**Data processing tools**

Standard pH meter, temperature, D.O, conductivity, B.O.D. and Nitrate measurement instruments were used at the time of collecting data. For processing the data, we used Microsoft Excel. A core i5 laptop PC was used. The PC was configured with 2GB NVIDIA graphics memory with 2GB Intel graphics memory, 12 GB RAM, and 1TB SSD.

**Experimental setup and training**

The GBM (Gradient Boosting Model) library was installed and called to train the Gradient Boosting Model in R [29-31]. Specific arguments and calculations are needed for functioning the GBM. The necessary arguments and calculations are mentioned in the method. Using the arguments, the model generates predictor variables. Then the response variables are specified. If it is not defined, the model guesses a value. Finally, the data is determined. As GBM is an ensemble of decision trees, the n-trees arguments are also specified. Then the training process gets started, and after training and testing, the model provides results and predictions. The training cost has been mentioned in **Figure 4**. The loss function is minimized with the increasing number of iterations. The loss was at the peak initially, but it decreased when the training process forwarded. The X-axis of the graph shows the number of iterations, and Y-axis represents the cost.

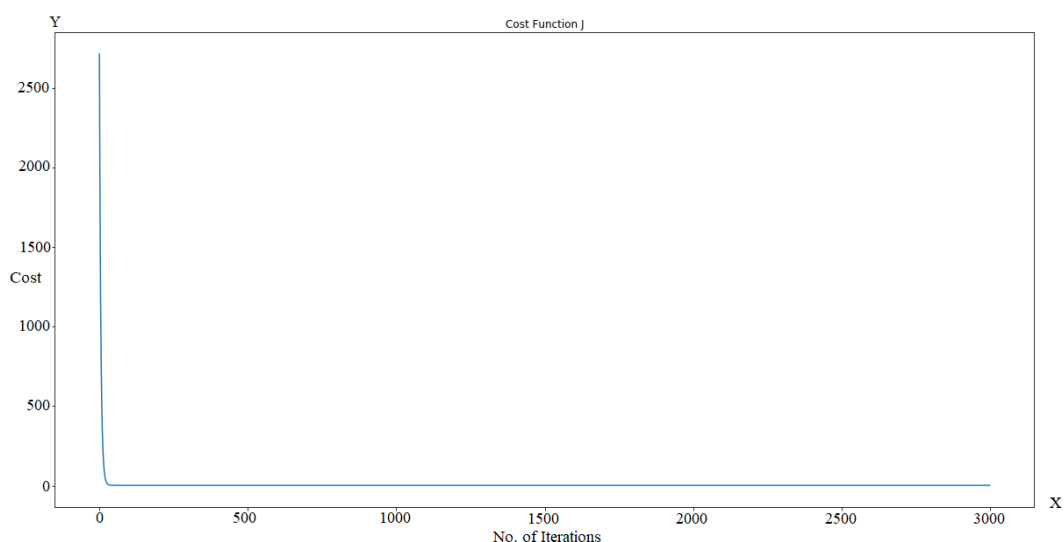

**Figure 4** Cost function.

**Result and discussion**

The poor learner is built up in each training round, and predictions are measured with the right outcome. The difference between prediction and reality represents the model's error rate. The errors were used to calculate the gradient, which was the loss function's partial derivative. The gradient performed direction-finding. Following the training and testing, the model provided detailed analysis shown in **Table 6**. The model yields prediction depending on the analyzed result. The model can predict the water quality and the diseases caused by specific water quality features. A plot of the actual result and predicted values are illustrated in **Figure 5**.

**Figure 5** has established a relationship among two factors, the data collection year and the WQI. The year, shown in the X-axis, is an independent variable, and the WQI, shown in the Y-axis, is a dependent variable. The straight line approximates the relationship between the two axes. The line slants upward with the lower stopping point at the Y-catch of the diagram and the upper stopping point broadening upward into the diagram field, away from the X-axis. The two variables exhibit a positive linear relationship. The model fits with data very well. It has a Mean Square Error of 1.1987755149740886, and R squared is 74.63 %, approximately 75 %. The standard deviation of the blunders is one portion of the standard deviation of the dependent variable.

**Table 6** Data analysis

| S.N | Station | Location | do | pH | co | bod | na | tc | year | … | nbdo | nec | nna | wph | wdo | wbdo | wec | wna | wco | wqi |
|---|---|---|---|---|---|---|---|---|---|---|---|---|---|---|---|---|---|---|---|---|
| 2 | 1208 | Mirpur Area, Dhaka | 6.3 | 6.9 | 179.0 | 1.7 | 0.1 | 5330 | 2019 | … | 100 | 60 | 100 | 13.2 | 28.10 | 23.40 | 0.54 | 2.8 | 11.24 | 79.28 |
| 3 | 9320 | Dighala Area, Khulna | 5.8 | 6.9 | 64.0 | 3.8 | 0.5 | 84443 | 2019 | … | 80 | 100 | 100 | 13.2 | 22.48 | 18.72 | 0.90 | 2.8 | 11.24 | 69.34 |
| 4 | 9140 | Tala Area, Satkhira | 6.1 | 6.7 | 308 | 1.4 | 0.3 | 5672 | 2019 | … | 100 | 0 | 100 | 9.9 | 28.10 | 23.40 | 0.00 | 2.8 | 11.24 | 75.44 |
| … | … | … | … | … | … | … | … | … | … | … | … | … | … | … | … | … | … | … | … | … |
| 1997 | 2614 | Jessore Lake Area | 4.6 | 3.0 | 350 | 6.2 | 2.2 | 49.0 | 2018 | … | 60 | 0 | 100 | 0.0 | 16.86 | 14.04 | 0.00 | 2.8 | 22.48 | 56.18 |
| 1998 | 2623 | Magura Area | 10.00 | 7.1 | 150.0 | 1.0 | 4.0 | 35. | 2018 | … | 100 | 80 | 100 | 16.5 | 28.10 | 23.40 | 0.72 | 2.8 | 16.86 | 88.38 |
| 2000 | 3010 | Foridpur Area | 9.00 | 7.3 | 158 | 1.8 | 7.2 | 280.0 | 2018 | … | 100 | 60 | 100 | 16.5 | 28.10 | 23.40 | 0.54 | 2.8 | 16.86 | 88.20 |

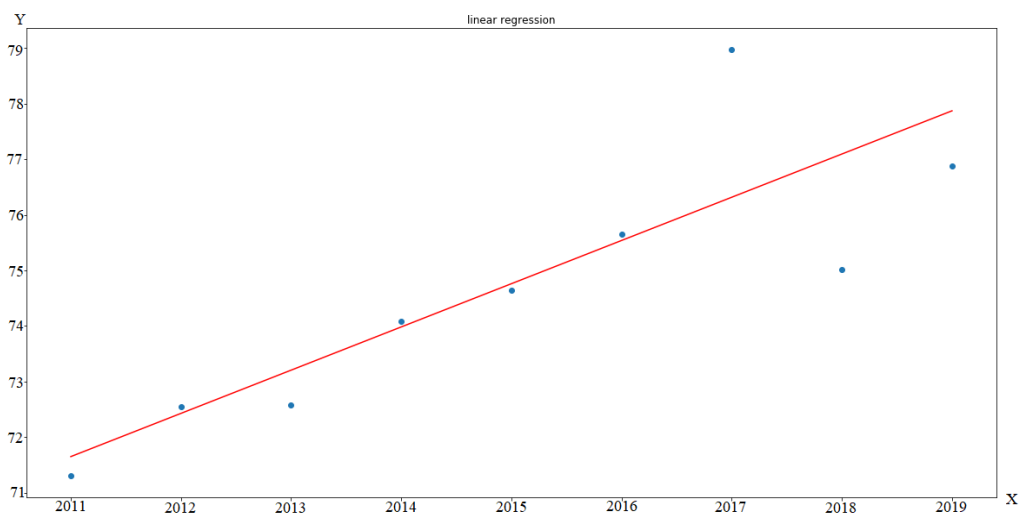

**Figure 5** Plotting the actual and predicted results.

**Table 7** shows the predicted water quality, prediction errors, the effect on fish due to the change of WQI, and the intelligent decision prepared by the model. The column "WQI" shows the present quality of a data. The "Predicted WQI" column shows the prediction of the previous "WQI" columns data. We collected samples from 500 different sources. From each source, we collected three samples by four months intervals and trained them to the model. The model predicts four months' future by analyzing the given data. The "Percentile error" column judges the accuracy of the prediction. The model is trained with the possible diseases caused by the change of the WQI score, and it takes intelligent decisions against it. The column "Chances of disease" and "Decision" show the machine-generated output for the current change of WQI. The decision can be effective against the fish disease. The model can predict WQI and measure diseases with an average of 92 % accuracy.

Table 7 Prediction, diseases identification and smart decision.

| Serial | WQI | Predicted WQI (after 4 months) | Percentile error | Chances of diseases | Decision |
|---|---|---|---|---|---|
| 1 | 63.253922 | 69.959334 | 10.6 % | No production | Minimize acidity by using soda lime |
| 2 | 78.969041 | 83.966075 | 6.3 % | No disease | Comfortable |
| 3 | 77.549000 | 81.307586 | 4.8 % | No disease | Comfortable |
| 4 | 75.058490 | 67.314328 | 10.4 % | Slow growth | Protein synthesis |
| 5 | 50.570943 | 52.655839 | 4.1 % | White sturgeon | Use potassium |

**Conclusions**

This work presents a machine learning-based automated solution for identifying fish diseases based on chemical contents analysis in water habitats. The proposed model incorporates a sophisticated classifier algorithm that can learn a real dataset to assess water quality and eventually to infer intelligent decisions. One significant aspect of the research is collecting and processing the dataset used to benchmark the proposed model performance. The model can produce a realistic decision on actions against potential diseases in the habitat. The accuracy of the model is found to be very satisfactory.

The early warning for the fish diseases will undoubtedly be of great relief to the fish farmers as this model not only help to detect disease but is also capable of referring to effective preemptive action to avert the disease. Furthermore, this study positions itself on a great potential to explore further extension to incorporate more types of diseases and mitigating decisions. IoT-based implementation should be analyzed based on the proposed algorithm to equip fish farmers with the necessary technology for quick and reliable on-site usage.